\documentclass{svproc}
\usepackage{url}
\usepackage{graphicx}
\usepackage{float}
\usepackage{supertabular}

\usepackage{amsmath}
\usepackage{amsthm}
\usepackage{amssymb}
\usepackage{longtable}
\usepackage[numbers]{natbib}
\usepackage{natbib}

\begin{document}
\mainmatter              % start of a contribution
%------------------------------------------------------------------------
\title{Supporting Financial Inclusion with Graph Machine Learning and Super-App Alternative Data}

% Enhancing Credit Scoring with Graph Machine Learning in a Super-App
% Enhancing Credit Scoring using Super-App Alternative Data with Graph Machine Learning
% A Graph Machine Learning Approach to Credit Scoring using Super-App Alternative Data 

% CAMBIO. enfoque en loq eu pusimos en el ultimo parrafo

% Supporting/Enhancing Financial Inclusion with Graph Machine Learning using Super-App Alternative Data
%

\titlerunning{Credit Scoring using Graph Machine Learning}  % 

\author{Luisa Roa\inst{1} \and Andr\'es Rodr\'iguez-Rey\inst{2} \and Alejandro Correa-Bahnsen\inst{1} \and Carlos Valencia Arboleda \inst{3}}

\authorrunning{Roa et al.} % abbreviated author list (for running head)

%%%% list of authors for the TOC (use if author list has to be modified)
\tocauthor{Luisa Roa, Andr\'es Rodr\'iguez-Rey, Alejandro Correa-Bahnsen, Carlos Valencia Arboleda}

\institute{
Rappi, Bogot\'a, Colombia \\
\and 
University of California, San Diego, La Jolla, CA\\
\and
Universidad de los Andes, Bogot\'a, Colombia
}

\maketitle              % typeset the title of the contribution

\let\thefootnote\relax\footnote{
\scriptsize NOTICE: this is the author's version of a work accepted for publication. Changes resulting from the publishing process, such as editing, corrections, structural formatting, and other quality control mechanisms may not be reflected in this document. Changes may have been made to this work since it was submitted for publication.

\textit{Email Addresses}: \email{luisa.roa@rappi.com} (Luisa Roa), \email{a3rodrig@ucsd.edu} (Andr\'es Rodr\'iguez-Rey), \email{alejandro.correa@rappi.com} (Alejandro Correa-Bahnsen), \email{cf.valencia@uniandes.edu.co} (Carlos Valencia Arboleda)
}

\begin{abstract}
The presence of Super-Apps have changed the way we think about the interactions between users and commerce. It then comes as no surprise that it is also redefining the way banking is done. The paper investigates how different interactions between users within a Super-App provide a new source of information to predict borrower behavior. To this end, two experiments with different graph-based methodologies are proposed, the first uses graph based features as input in a classification model and the second uses graph neural networks. Our results show that variables of centrality, behavior of neighboring users and transactionality of a user constituted new forms of knowledge that enhance statistical and financial performance of credit risk models. Furthermore, opportunities are identified for Super-Apps to redefine the definition of credit risk by contemplating all the environment that their platforms entail, leading to a more inclusive financial system.

\end{abstract}

\section{Introduction}
 
%Rappi is the largest Super-App in Latin America, present in 9 countries and with more than 50 million users, it offers a wide range of services such as food delivery, errands, booking of flights and hotel rooms, video games, a gambling platform,  and  financial services through a virtual wallet that allows P2P transfers, bank transfers, money withdrawals, and payments with QR. With the inclusion of new financial services, specifically consumer loans, the importance of accurate descriptions of credit risk and loan allocation becomes crucial for the company, especially for credit risk management since it is necessary to identify users who are not suitable for a loan and represent a high risk of default.

Super-Apps are platforms that offer a wide range of services such as food delivery, errands, booking of flights and hotel rooms, video games, a gambling platform, and even financial services like P2P transfers, bank transfers, money withdrawals, and payments with QR. A clear example is WeChat from the Chinese giant Tencent, that started as a messaging application and now offers personal financing solutions through its Fintech WeBank \citep{DBS2019}. Given that these platforms have a high potential to become Fintechs, specifically focused on consumer loans, the importance of accurate descriptions of credit risk and loan allocation becomes crucial, especially for credit risk management since it is necessary to identify users who are not suitable for a loan and represent a high risk of default. %cita de que son muy grandes

To manage default risk, financial institutions implement credit scoring systems to assess the risk associated with each loan application \citep{MEDINA2013} and assign a credit score to clients that captures the estimated individual probability of default. In traditional financial entities, the credit score is based on the financial history of the individual with information that includes the number of credit lines, loan amount, installments, previous payments, and bureau score among others. This type of data is not only expensive and scarce, it also has the problem that relies solely on the financial profile of individuals, leaving anyone without a history of traditional banking products excluded from acquiring financial services. In Latin America and the Caribbean, for instance, there are more than 200 million unbanked adults (more than 10\% of the unbanked worldwide) \citep{Asli2018} who cannot access these services due to lack of financial information. That is why the main challenge is to find alternate metrics to evaluate individuals credit worthiness, in such a way that the metric of what we consider to be ``worthy" does not depend exclusively in their financial history.

As an alternative, some authors have considered the data generated by users in different types of digital platforms \citep{berg2020rise}. In particular, the data degenerated by users in the different functionalities of Super-App's have shown to capture valuable information for credit evaluation that the financial history cannot capture~\citep{roa2021super}. From the latter, it becomes apparent that the quality of the transactional data that a Super-App captures is different to that a traditional bank may record. Moreover, due to the nature of these platforms and the interactions between users and different entities such as stores, devices, products, and others, a new type of information is captured. The ecosystem generated by these interactions allow the construction of networks within the app, which provides better understanding of the users. These graphs constitute an advantage for Super-Apps, since, as it has been recently showed, these type of alternative data capture highly predictive information for problems involving fraud detection~\citep{chen2020infdetect,liu2020heterogeneous}, credit assessment \citep{oskarsdottir2019value}, recommendation systems \citep{xiang2010temporal}, among others.

There is a new trend of research that pushes for the use of these networks and the information they capture to contribute in the assessment of credit worthiness. For instance, \citet{westland2018private} showed that the topology of a network based on communications and travel information from borrowers becomes relevant as it  allows to improve the credit score between peers and their risk assessment. The authors identify that users who have a dense network are less likely to default. Along the same lines, \citet{bravo2020evolution} define a multi-layered network in which they study the correlated default using the personalized PageRank algorithm. The authors show that by using network theory, the propagation of default on networks can be measured and used to provide valuable information for credit assessment as network centrality scores show trends related to the default rate. Furthermore, several authors have proposed methods that take advantage of the structure of the graphs directly in order to predict the traditional credit scores. \citet{jalilifard2020friendship} builds a friendship network and proposes a multi-graph embedding approach that seeks to predict traditional banking creditworthiness through the behavior of groups of close friends. Jalilifard et al. found that when using the embedding as input to a logistic regression the performance is comparable with state of the art methods, suggesting that this could be an alternative for computing traditional banking credit scores. On the other hand, \citet{lin2018netdp} proposed a network representation framework for default prediction in which a Multiple Additive Regression Tree is used to predict the probability of default when assembling a supervised and an unsupervised module, where the first models the local structural information while the second focuses on global structural information. By creating a network of user interactions, they demonstrate that the model based on graphs exceeds the performance of the credit score prediction for new users on the platform and strengthens the benchmark models for old users. 

The objective of this paper is to examine the statistical impact of graph based features and models in traditional credit risk assessment and their financial impact on the company lending profitability, as well as to motivate the need of a new way of defining credit worthiness. For this, we will construct different networks based on 5 types of user interactions within the super app: P2P transfers, relationships due to having the same card, the same device, sharing the same geolocation based on the delivery addresses, and the first six digits of a card known as BIN. From these networks, graph-based features will first be obtained and then used as input in a classification model to evaluate the statistical and financial gain that each type of interaction represents in credit risk. Then, for one of these networks, the effect of using graph neural networks as a model is explored. It is important to note that in this paper no financial information from bureaus or financial allies is considered in the construction of the features, and because there is still no financial product from the Super-App on the market a financial partner will provide the data of those users who have fallen into default. 

The remainder of the paper is structured as follows: Section 2, presents the technical details of the experimental methods used in the paper; Section 3, describes the experiments, a detailed account of the data, as well as the results; Section 4, presents the results; Section 5, discuss the financial implications arise from results and, finally, in Section 6, we hold the conclusions and future directions.

\section{Preliminaries}

In this section we give some details on the technical aspects of the methods we are going to use on our experiments.

\subsection{XGBoost}

Extreme gradient boosting, or best known as XGBoost, is a scalable supervised machine learning algorithm for tree boosting \citet{chen2016xgboost}. The algorithm is based on the use of different techniques to improve performance and execution speed. It starts from the concept of boosting, an ensemble method in which the models, in this case decision trees, are built sequentially in order to learn from errors to enhance the prediction. Likewise, as its name implies, it uses gradient descent to calculate errors or residuals from an objective function that measures the difference between the prediction and the real value. With the above, along code optimization and algorithmic enhancements, the XGBoost has become one of the algorithms with the highest prediction effectiveness.

\subsection{Graph Neural Networks}
A key problem in Geometric Deep Learning (GDL), the branch of machine learning that tries to generalize deep neural networks to data with non-euclidean structure, is that of classification. In this subsection we hope to give a high level introduction to the semi-supervised/supervised methods in GDL used in this paper. We start by defining a graph $G= (V,A, X, L)$ to be a set of nodes $V$ with adjacency matrix $A$, together with a matrix of node features $X$, and a set of node labels $Y$. We recall that $(A)_{ij}$ corresponds to the weight of the arc going from the $i$-th vertex to the $j$-th vertex, that $D$ is the diagonal where $(D)_{i,i}$ is the degree of the $i$-th vertex, and notice that $X$ is an $n\times |V|$ matrix where $n$ is the number of features and $|V|$ the number of vertices, moreover the $i$-th column of $X$ represents the vector of features of the $i$-th vertex of $G$. We also consider the set $\mathcal{N}_i$ of neighbors of the vertex indexed by $i$. It should be noticed that the number of nodes with a label $m$ might be less than $|V|$, in which case the problem is semi-supervised in nature, and the labels can be taken from a set $\{1,\dots, c\}$.

\subsubsection{Graph Convolutional Neural Networks}
Graph convolutional networks, defined by Kipf et al. in \cite{GCN}, are multi-layer feedforward neural networks that propagate and transform node features across a graph, its propagation rule is given by 
\[
X^{(k+1)}= \sigma \left( D^{-\frac{1}{2}}\tilde{A}D^{-\frac{1}{2}}X^{(k)}W^{(k)}\right),
\]
where $W^{(k)}$ is a trainable weight matrix in the $k$-th layer, $\sigma(\cdot)$ is an activation function, $\tilde{A}= A+I$, and $X^{(k)}$ is the $k$-th layer node representation where $X^{(0)}=X$. 

\subsubsection{GraphSage}
GraphSage is a non-spectral approach  that computes node representations in an inductive manner. It was introduced by \citeauthor{GraphSage}, as an alternative to many of the transductive methods in the literature. Graphsage considers the neighborhood of each node, and then performs a specific aggregator over it, the result then fed into a recurrent neural network in order to propagate information between different layers of the model. This can be expressed as
\begin{align*}
    x^{(k+1)}_{\mathcal{N}_i}&= \text{aggregate}\left(\{ x^{(l)}_j: j\in \mathcal{N}_i\} \right)\\
    \tilde{x}^{(k+1)}_{i}&=\sigma\left(W\cdot \text{concat}(x^{(k)}_i,x^{(k+1)}_{\mathcal{N}_i}) \right)\\
    x^{(k+1)}_{i}&=\text{norm}(\tilde{x}^{(k+1)}_{i}).
\end{align*}
In practice the aggregator is taken to be the average, and the normalizer to be normalization with respet to the $l^2$ norm.

\subsubsection{Graph Attention Network}
Graph Attention Networks (GAN) were introduced by \citeauthor{GAN} in \cite{GAN}, as an attention-based architecture to perform node classification of graph-structured data. It consists of a stack of \textit{graph attentional layers} defined by the rule
\[
x^{(k+1)}_i= \sum_{j\in \mathcal{N}_i} \alpha_{ij} W^{(k)}x^{(k)}_j
\]
where the $\alpha_{ij}$ are normalized attention coefficients, and $W^k$ is the corresponding input linear transformation's weight matrix. More explicitly, we have that  
\[
\alpha_{ij} = \text{softmax}_j(e_{ij})= \frac{\exp(e_{ij})}{\sum_{k\in \mathcal{N}_i}\exp(e_{ik})}
\]

where $e_{i,j}$ is the attention coefficient that indicates the importance of node $j$'s features to node $i$ in terms of a shared attentional mechanism $a$, that is,
\[
e_{ij}= a\left(Wx_i, Wx_j\right).
\]
 %TO DO
\subsubsection{TAGCN}
Topology Adaptive Graph Convolutional Networks (TAGCN) were introduced by \citeauthor{TAGCN} in \cite{TAGCN}, these networks are a vertex domain approach to the convolutional neural network problem on graphs where the graph convolution operation is defined in terms of a shift-invariant filter, this filter can be expressed as as a multiplication by polynomials of the graph adjacency matrix. More explicitly, 
\[
x^{(k+1)}= \sigma\left( \sum_{c=1}^{C_k} \sum_{j=0}^K g^{(k)}_{c,j} \left(D^{\frac{-1}{2}}AD^{\frac{1}{2}}\right)^j x_c^{(k)}\right),
\]
where $\sigma$ is a ReLU activation function, $C_k$ is the dimension of the $k$-th layer, $K$ is the degree of the filter as a polynomial of the adjacency matrix, and the $g^{(k)}_{c,j}$ are the coefficients of said polynomial. The reader should notice that $x_c^{(k)}$ is the vector corresponding to the entries of $c$-th feature of each vertex in the $k$-th layer of the network. 

\subsection{Financial evaluation} 

Although the statistical evaluation of a model is essential to know its effectiveness, in problems such as credit risk assessment it is also necessary to evaluate the model with some performance measure that considers business costs in order to assess the profitability generated by the model. \citet{CorreaBahnsen2014b} presented a cost sensitive approach for credit risk assessment where the cost of a wrong prediction is not constant among the observations and is evaluated as presented in Table~\ref{tab:c_mat}.

\begin{table}[h]
\centering
\caption{Credit scoring example-dependent cost matrix}
\label{tab:c_mat}
\begin{tabular}{lcc} 
\hline
       & \multicolumn{1}{l}{Actual Positive} & \multicolumn{1}{l}{Actual Negative}  \\ [2pt]
\hline
Predicted Positive     & $C_{TP_i}=0$                   & $C_{FP_i}=r_i+C^a_{FP}$\\
Predicted Negative     & $C_{FN_i}=Cl_i \cdot L_{gd}$   & $C_{TN_i}=0$\\[2pt]
\hline
\end{tabular}
\end{table}

%-------------------------------------
%\begin{table}[h]
%\caption{This is the example table taken out of {\it The
%\TeX{}book,} p.\,246}
%\begin{center}
%\begin{tabular}{r@{\quad}rl}
%\hline
%\multicolumn{1}{l}{\rule{0pt}{12pt}
%                   Year}&\multicolumn{2}{l}{World population}\\[2pt]
%\hline\rule{0pt}{12pt}
%8000 B.C.  &     5,000,000& \\
%  50 A.D.  &   200,000,000& \\
%1650 A.D.  &   500,000,000& \\
%1945 A.D.  & 2,300,000,000& \\
%1980 A.D.  & 4,400,000,000& \\[2pt]
%\hline
%\end{tabular}
%\end{center}
%\end{table}
%--------------------------

The cost of correct predictions is zero while the cost of a false positive $C_{FP_i}$ is the loss over the credit line define as the credit line $Cl_i$ times the loss given default $L_{dg}$ and the cost of a false negative $C_{FP_i}$ is the sum of the profit $r_i$ when declining a good payer plus the cost of giving the credit to another costumer $C^a_{FP}$. Once the cost-matrix is defined, the cost improvement can be expressed as the cost savings as 
  $$    Savings = \frac{ Cost_l - Cost}   {Cost_l},$$  
where $Cost$ is calculated as the sum of individual costs matrices for each customer
  $$   Cost = \sum (1-c_i)*y_i*C_{FN_i} + (1-y_i)*c_i*C_{FP_i},$$
and $Cost_l$ is the cost of the cost-less class overall results of a model. The financial performance of the experiments in this paper will be evaluate with the savings metric previously presented in order to identify those models that provide greater profitability.

% \section{Experiments and Results}

% In this section, we present details on the constraints of our data,  a summary of the variables we use, the experimental setup, as well as the results of said experiments.

\section{Credit Scoring using Graph Machine Learning}

As mentioned above, one of the biggest struggles in credit risk scoring is the lack of means to access financial data, this includes credit scores for training purposes. Because all of the credit risk information available for this paper was given to us by a financial ally of the Super-App in the country of study, the whole population was taken from a sample of those Super-App users who had a connection to the ally. For this population we then computed some Super-App market based features that hoped to model financial aptitude, as well as graph based data that captured social interactions or similarities between users. The data set used in the experiments consists of a sample of 38,342 of a Latin American Super-App who have defaulted on a credit product of the ally and the total default rate is 12.9\%. For each user, 149 Supper-App and 11 graph-based features per graph are considered. Among the Super-App features, we consider data that captures demographic, behavioral, and transactional information about the user.

%graphs images-----------------------
\begin{figure}[tbp!]
\centering
 \includegraphics[width=0.49\linewidth, height=5.5cm]{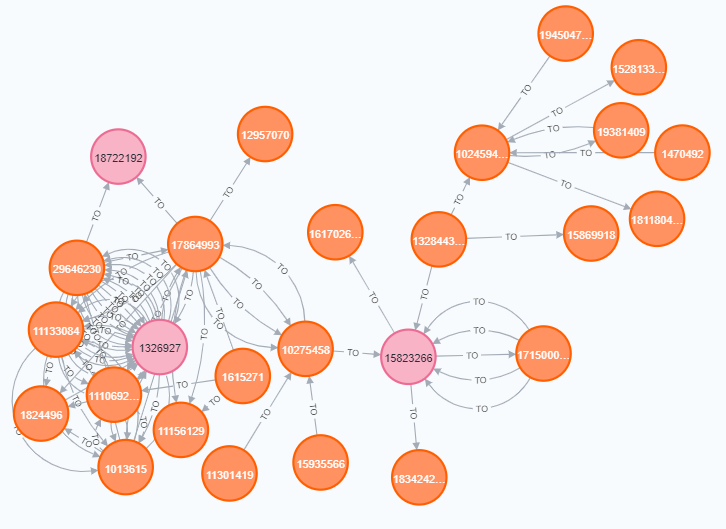}
 \includegraphics[width=0.49\linewidth, height=5.5cm]{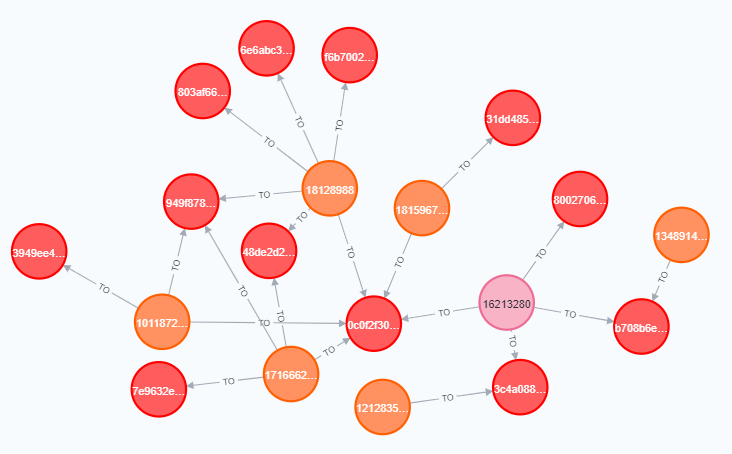}
\caption{The first graph represents the P2P network and the second the CC network. Users are represented by pink (defaulters) and orange nodes (non-defaulters), credit cards are shown in red.}
\label{fig:graph_ex}
\end{figure}

In hopes of fully taking advantage of the possibilities of the Super-App data, we decided to consider different networks that captured the behavior of the users inside the Super-App: 
\begin{enumerate}
    \item \textbf{P2P}: The graph of peer-to-peer transactions in the virtual wallet platform of the Super-App. This graph is a directed graph on the set of users, where for each edge its source is the user who initiated the transfer, and its direction points towards the user who receives the transfer. When considering all of the interactions of 38,342 users the network ends up having 88,270 vertices and 214,637 edges.
    \item \textbf{BIN}: In hopes to better understand users with similar socioeconomic status we decided to construct a graph that allowed us to capture information from users with similar financial products. We decided then to construct the graph that relates users to the BINs, where we only kept those users and BINs inside the population of interest and those users that used any of said BINs at least once in October 2020. The graph has as nodes the BINs and the users, and the graph's edges are directed edges from users to BINs. The graph has 901,366 users, 9,096 BINs, and 1,646,201 edges. 
    \item \textbf{GEO}: To capture users with similar sociodemographic data we decided to construct a graph whose vertices consisted of geohashes and users, and whose edges connected users with the geohashes from the delivery addresses used inside of the App. We then, as with the BIN-graph, restricted the graph to those users and geohashes related to our population of interest together with the users who had been in said geohashes at least once in October 2020. This graph has 276,260 users y 34,224 geohashes and 1,104,142 edges.
    \item \textbf{CC}: In order to capture users with similar financial background we considered the graph that consists of users and credit cards, where the edges connect a user with one of their registered credit cards. Inside of a Super-App it is not uncommon for different users to have registered the same credit cards, in order to consider the financial behavior of those users that may share payment methods with the population given to us by the Super-App ally, we decided to inductively grow our network. The graph ended up having 136,009 users, 385,014 credit cards and 634,870 edges.    
    \item \textbf{DV}: Finally, we construct a graph whose vertices consist of users and devices, and where the edges relate users with their registered devices. The graph was constructed in a similar way to the CC-graph. This graph has 247,844 users, 385,014 devices and 707,948 edges. 
\end{enumerate}

%graphs images-----------------------
\begin{figure}[tbp!]
\centering
 \includegraphics[width=0.49\linewidth, height=5.5cm]{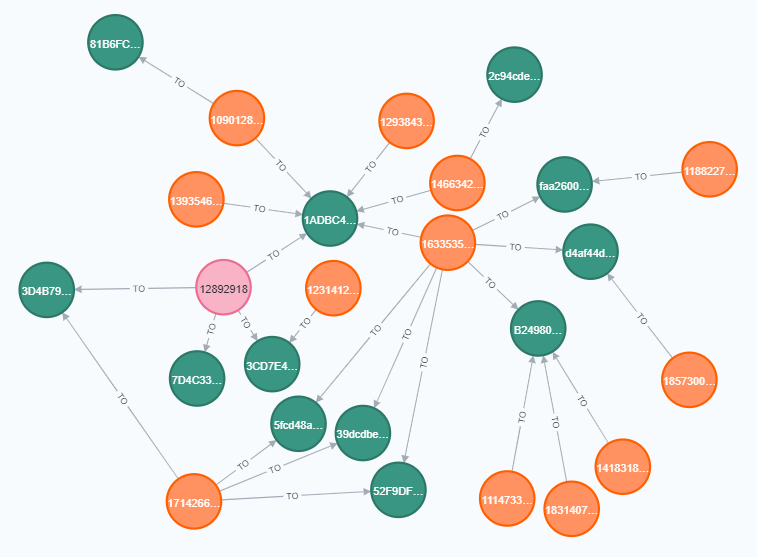}
 \includegraphics[width=0.49\linewidth, height=5.5cm]{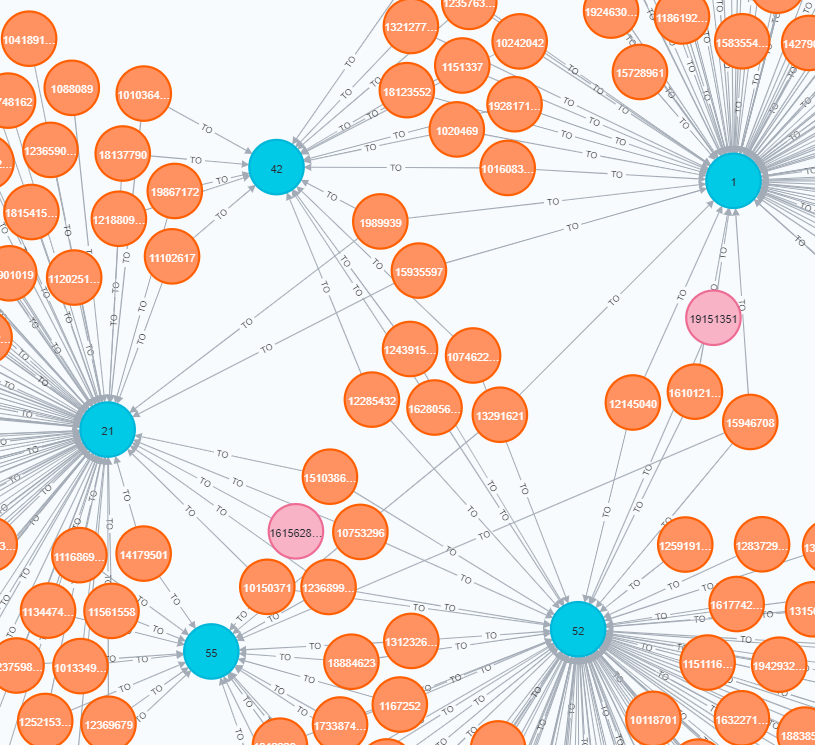}
% \medskip
% \includegraphics[width=0.49\linewidth, height=5.5cm]{images/geo_d.PNG}
\caption{The first graph represents the DV network and the second the Bins. Users are represented by pink (defaulters) and orange nodes (non-defaulters),  devices and BINs are shown in green and blue.}
\label{fig:graph_ex2}
\end{figure}

%Graphs structure details table ------------
\begin{table}[h]
\begin{center}
  \caption{Graphs structure details}
  \label{tab:nodes_edeges}
  \begin{tabular}{ccc}
    \hline
    Graph&Nodes&Edges\\
    \hline
    P2P         & 88,270    & 214,637\\
    Credit Card & 576,042   & 634,870\\
    Devices     & 632,858   & 707,948\\
    Bines       & 910,431   & 1,646,201\\
    Geohash     & 310,484   & 1,104,142\\
  \hline
\end{tabular}
\end{center}
\end{table}

A summary of these graphs is on Table \ref{tab:nodes_edeges} and as an example, Figures \ref{fig:graph_ex} and \ref{fig:graph_ex2} present some relationships and default users. From these networks, we computed what we call \textit{graph-features}, which include centrality, community, and neighborhood variables, for each of the users the population of interest:

\begin{enumerate}
    \item \textit{Total degree}: Total number of connections of a node.
    \item \textit{Eigenvector centrality}: Score for each node based on the importance of its neighbors, for directed graphs is obtain by finding  the right leading eigenvector $x$ that satisfies $Ax=\lambda x$ where A is the adjacency matrix and $\lambda$ is the largest eigenvalue of A.
    \item \textit{Pagerank}: Score that measures the importance of a node by randomly traversing the graph and counting incoming and outgoing edges each node weighted by their importance.  
    \item \textit{Louvain}: It is a method that extract communities inside a graph and designates which nodes belong to said communities.
    \item \textit{Average neighbors features}: In this category, for each user we compute the averages of certain variables of the Super-App over their neighborhood. For instance, in the P2P network for user A that is connected to users B and C, an example of one type of neighborhood variable of A is the average of the orders cancelled with credit card of users B and C. It is considered important to obtain information from the neighborhood as it allows to capture information on the possible behavior of the user from their peers. The variables selected are: orders canceled due to payment error, orders paid with credit card, maximum credit card score, and if the users is prime.
\end{enumerate}

We then performed a set of experiments in order to train a classifier that could predict the risk of default in our population. These experiments can be divided into two types:
\begin{enumerate}
    \item \textit{Tabular}: For the tabular methods we used an XGBoost with 5 randomized bootstrap, trained with 70\% of the data and tested with the remaining 30\%, where we considered for each user their respective Super-App features and some, or none, of the graph-features.  We repeated this experiment six different times, one for each graph, where in each of them one considers the Super-App and the respective graph-based features of the graphs presented above; on top of one with all the graph-features we collected for then compare them to a base modeled that only considers Super-App features.

    \item \textit{GNN}: We then decided to run different types of semi supervised graph neural networks in the P2P graph considering only numerical variables of the Super-App. For each of the experiments the same structure of GNNs is used, only the type of layer and the additional parameters that each require, change. In general all GNNs consist of 2 layers, where our hidden layer has a ReLU activation function and 16 neurons and the last layer has an output size of 2. We train our models with respect to a cross entropy loss function, the learning rate is 0.02 and are trained with 200 epochs. Given that the data is imbalanced we decided to weight the cross entropy loss so that we could improve the precision of our models.  
\end{enumerate}

Finally, to evaluate the models statistical performance, we use the Area Under the ROC Curve (AUC) as it provides an aggregate measure which quantifies the ability to discriminate, in this case, a good payer from a bad payer across the entire range of possible thresholds. 

\section{Results}

\begin{figure}[h]
  \centering
  \includegraphics[scale = 0.60]{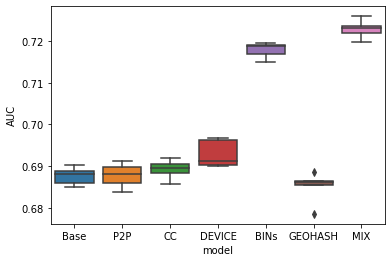}
  \caption{AUC performance by Tabular experiment}
  \label{fig:AUC}
\end{figure}

For the Tabular experiment the base XGBoost model, which only considers transactional and behavioral variables of the user, an average performance of 0.687 of AUC was observed. As it can be seen in Figure \ref{fig:AUC}, the models that consider variables based on graphs with relationships of  P2P transfers, credit cards, devices, and geohashes do not represent gain in terms of AUC, however the interactions through BINs manage to capture information that improves the predictive capacity by having a gain of the AUC of 0.030. In addition, when considering all graph-based variables within a single model, the greatest gain is found.  

\begin{figure}[h]
  \centering
  \includegraphics[scale = 0.60]{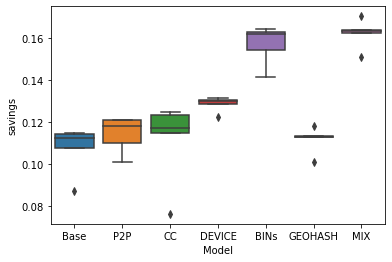}
  \caption{Savings by Tabular experiment}
  \label{fig:savings_xgboost}
\end{figure}

Furthermore, Figure \ref{fig:savings_xgboost} shows the average savings that each model represent. Overall, the results are similar to the statistical performance as the higher savings are provided by the model that considers all the features based on graphs, the gain compared to the base model is 5.5 percentage points. Additionally, the models with only P2P transfers, credit cards, and geohashes features represents the least financial improvement as savings increases in less than a percentage point respect to the base model. 

\begin{figure*}[h]
  \centering
  \includegraphics[scale = 0.40]{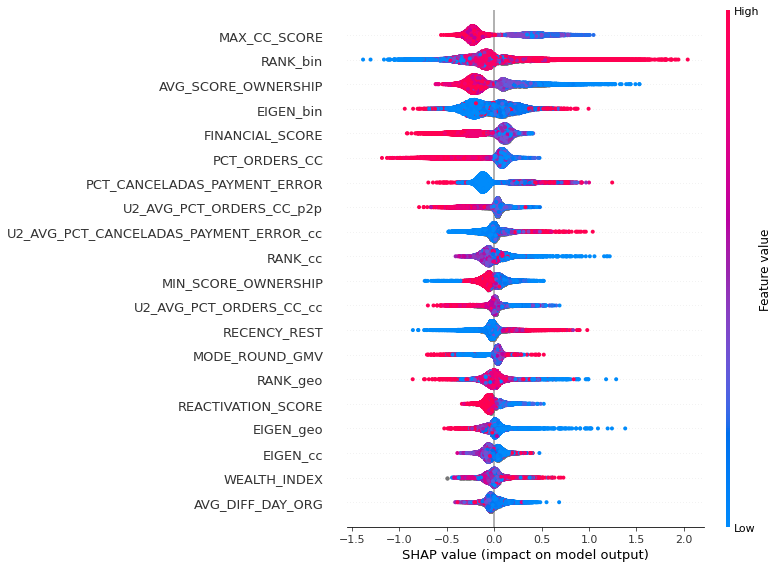}
  \caption{Feature importance and variables effects with SHAP}
  \label{fig:shap}
\end{figure*}

Since a black box model is used, we decided to use the Shapely Additive Explanations (SHAP) \citep{NIPS2017_7062} technique to understand the importance and effect of the variables of the graphs in the default prediction. In Figure~\ref{fig:shap} we present the \textit{summary plot} for the model that combines all the graph-based and Supper-App features. In this Figure the features are ordered according to their importance on the y-axis, and in the x-axis the variable's impact on the model output is recorded, where the colors represent the variable value: red is a high value and blue is low. The effects of each variable are interpreted in the following way: if the variable has high values in the positive part of the x-axis, that is, points with red colors, then a high value of the said variable increases the probability of being defaulter and low values decrease it. With this in mind, the two Super-App variables with the highest predictive value are the maximum credit card score and the average ownership score of registered cards, these have the same impact on the model output as a high value implies a lower probability of default. These variables behave as expected since they capture information related to banking history as well as any categorization given by a financial entity based on the card level.

%%%----y sin la explicación de shap?-------
%Since a black box model is used, we decided to use the Shapely Additive Explanations (SHAP) \citep{NIPS2017_7062} technique to understand the importance and effect of the variables of the graphs in the default prediction. In Figure \ref{fig:shap} we present the \textit{summary plot} for the model that combines all the graph-based and Supper-App features, this shows that 9 of the 20 most predictive variables for the default come from graphs. The two Super-App variables with the highest predictive value are the maximum credit card score and the average ownership score of registered cards, these have the same impact on the model output as a high value implies a lower probability of default. These variables behave as expected since they capture information related to banking history as well as any categorization given by a financial entity based on the card level.

Regarding the variables based on graphs, even though the most predictive variables come from the BINs graph, it can be identified that variables of the credit card, geohash and, P2P graphs manage to capture relevant information for the credit assessment even when in the individual  models there is no significant gain. The centrality measure by the pagerank algorithm in the graph of BINs is the second most predictive variable and the greater the measure the greater the probability of default, although the eigenvector centrality also seeks to capture the importance in the graph, it is evident that this variable seems to capture a non-linear relationship with the default as the high (red) and low (blue) values are distributed both the positive and negative side of the SHAP values in Figure~\ref{fig:shap}. The fact that these variables are highly predictive indicates that BINs manage to group users with a similar financial profile, even if the card is not their own, consequently a proxy for the economic capacity to acquire a loan and pay it is captured.

Other variables that show interesting results are those that obtain information from the neighborhood. First, as the neighbors of the P2P network has a higher percentage of orders canceled with a credit card, the user is less likely to default. This implies that the level of banking and the frequency of use of the credit card of the direct neighbors allow to infer about the creditworthiness. Similarly, an user has a greater probability of default if their neighbors on the credit card network have a high percentage of orders with payment error, since declined transactions may indicate liquidity problems or even abnormal behavior. 

Moreover, the performance results of each GNN are shown in Table \ref{tab:auc_gnn}. TAGCN achieves the best performance with an AUC of 0.65 while GAN shows to have a significantly lower predictive capability compared to all models, including the XGBoost experiments.

\begin{table}[h]
\begin{center}
  \caption{GNNs performance}
  \label{tab:auc_gnn}
  \begin{tabular}{ccl}
    \hline %\toprule
    GNN&AUC    \\
    \hline %\midrule
    GraphSAGE  & 0.62656 \\
    GCN        & 0.63578 \\
    TAGCN      & 0.65815 \\
    GAN        & 0.52825 \\
  \hline %\bottomrule
\end{tabular}
\end{center}
\end{table}

\section{Implications on Financial Inclusion}

In this section we discuss the impact of the results on financial inclusion, of utter importance when considering populations of unbanked adults and young individuals with no banking history but increasing interest in financial products from alternative lenders. The importance of new sources of information becomes necessary for the evaluation of risk for these populations.

Our results show that Super-App data that arises from BINs, credit card, P2P, and geohashes graphs is particularly relevant for credit assessment. This can be of particular importance when evaluating users with banking history who have little transactionality inside the Super-App. In contrast, the impact of variables like recency in restaurants, mode of gross margin value spend, and days between an organic order demonstrates how transactionality and preferences within the app provide value when assessing credit worthiness. The same holds true for variables related to the centrality of these graphs, which have similar importance to the data involving financial products. 
From this we can foresee that alternative data sources from Super-Apps  provide new ways to estimate the probability of default for an user, as well as a motivation for the opportunity that these Super-Apps face to better support financial institutions in methods that generate inclusion and decrease the costs of credit scoring.

Therefore, these Super-Apps should not limit their credit risk assessment to traditional financial data. Instead, they have the opportunity to redefine credit risk and create their own credit scoring that evaluates the reliability of an user based not only on payments of financial obligations but on alternative data that does not discriminate the underbanked population. We hope this paper pushes into that direction, which in countries of Latin America for example, will definitely lead to a more inclusive and sustainable credit system.

\section{Conclusion}

In this paper we studied the statistical and financial effects of considering variables based on graphs and the use of non-traditional models in the credit evaluation of users within a Super-App.

Overall, the XGBoost that considers all the variables outperform statistically and financially the different experiments presented. Furthermore, we find that variables derived from the BINs graph centrality seem to have a high effect in credit evaluation as well as centrality from credit card and geohashes and variables of the super-app related to the use of the credit card, use of discounts, and amount spent.  

On the other hand, the GNNs on the P2P network did not perform as expected. The poor performance could be explained by the size of the data set as well as the imbalanced nature of the data. In a larger network, with a more dense structure, the GNNs may be able to capture neighboring data in a way that was not reflected in this paper. While we believe that GNNs are a piece of technology that should be exploited in financial data this paper confirms the limitations of GNNs when dealing with imbalanced data sets, a common property of the data in the financial sector. 

Lastly, we discuss how alternative data sources from Super-Apps provide new ways to estimate credit risk using graph based machine learning algorithms, providing these platforms the opportunity to better support financial inclusion and at the same time decrease the costs of credit scoring.  

\section{Acknowledgments}

Some of the results in this paper were part of the Master thesis of the first author at Universidad de los Andes, under the direction of the last two authors. We would like to thank Jaime Acevedo, Gabriel Suarez and Juan R\'aful for their insightful ideas and valuable discussions.
 
%%\bibliographystyle{plainnat}
%%\renewcommand{\bibname}{References}
% \section*{References}

%%\bibliography{bib_paper}
%%\nocite*{}

% ---- Bibliography ----

\end{document}